\documentclass[sigconf]{acmart}

\usepackage[disable]{todonotes}

\usepackage{subcaption}
\AtBeginDocument{%
  \providecommand\BibTeX{{%
    \normalfont B\kern-0.5em{\scshape i\kern-0.25em b}\kern-0.8em\TeX}}}

\copyrightyear{2022}
\acmYear{2022}
\setcopyright{acmcopyright}\acmConference[GECCO '22]{Genetic and Evolutionary
Computation Conference}{July 9--13, 2022}{Boston, MA, USA}
\acmBooktitle{Genetic and Evolutionary Computation Conference (GECCO '22), July
9--13, 2022, Boston, MA, USA}
\acmPrice{15.00}
\acmDOI{10.1145/3512290.3528839}
\acmISBN{978-1-4503-9237-2/22/07}

\begin{document}

\title{Minimal Neural Network Models for Permutation Invariant Agents}


\author{Joachim Winther Pedersen}
\email{jwin@itu.dk}
\affiliation{%
  \institution{IT University of Copenhagen}
  \streetaddress{2300 Copenhagen, Denmark}
  \city{Copenhagen}
  \country{Denmark}
  \postcode{43017-6221}
}

\author{Sebastian Risi}
\email{sebr@itu.dk}
\affiliation{%
  \institution{IT University of Copenhagen}
  \streetaddress{}
  \city{Copenhagen}
  \country{Denmark}}
\email{}


\begin{abstract}

Organisms in nature have evolved to exhibit flexibility in face of changes to the environment and/or to themselves. Artificial neural networks (ANNs) have proven useful for controlling of artificial agents acting in environments. However, most ANN models used for reinforcement learning-type tasks have a rigid structure that does not allow for varying input sizes.\todo{Can organisms in nature deal with varying input sizes?} Further, they fail catastrophically if inputs are presented in an ordering unseen during optimization. We find that these two ANN  inflexibilities can be mitigated and their solutions are simple and highly related. 
For permutation invariance, no optimized parameters can be tied to a specific index of the input elements. For size invariance, inputs must be projected onto a common space that does not grow with the number of projections. Based on these restrictions, we construct a conceptually simple model that exhibit flexibility most ANNs lack. We demonstrate the model's properties on multiple control problems, and show that it can cope with even very rapid permutations of input indices,  as well as changes in input size. Ablation studies show that is possible to achieve these properties with simple feedforward structures, but that it is much easier to optimize recurrent structures.

\end{abstract}




\maketitle

\section{Introduction}
\label{section:intro}
Artificial neural networks (ANNs) have shown to be able to solve a wide array of challenging tasks \cite{schmidhuber2015deep}, including in the field of reinforcement learning and the development of artificial agents \cite{arulkumaran2017brief}. Consider for illustration a standard feedforward network (FFN). By tuning an often large amount of parameters, such a network can be made to approximate any function \cite{hornik1991approximation}. However, once the parameters have been optimized, they are tied to the specific structure that they were optimized in. This rigidity does most often not allow for adding elements to the input after training without having to optimize an entirely new neural network from scratch.

The fact that the parameters of an optimized ANN are strictly tied to their specific indices in the network architecture, also means that they are dependent on the elements of the external inputs always being presented in the same ordering. There are several downsides to this inflexibility. First, we might not always know beforehand the number of inputs that are ideal for our model. Ideally we would be able to add a new sensor and the model would during deployment figure out how to integrate that information. 
However, most neural architectures do not easily allow this, and we would be forced to start all over, if we wanted to incorporate the new input.

Additionally, there might be cases where we cannot guarantee that inputs to our model will always arrive in the same ordering as during the optimization phase. When parameters are tied to the indices of the ANN, any permutation of the input elements might be catastrophic.
Recently, multiple methods have been proposed to mitigate the inflexibility of ANNs and make them invariant to permutations of inputs \cite{tang2021sensory, kirsch2021introducing}. In these studies, it was also found that the properties of invariance to permutations and changes in size of the input vector was accompanied by extra robustness of the models to unseen perturbations after training.

This paper aims to contribute to this trend of making the neural architectures of artificial agents more flexible. We do so by proposing a conceptually simple model that after optimization can output coherent actions for a performing agent, even when the inputs to the model are continually shuffled in short intervals. Throughout, we emphasize the minimal requirements that an ANN must follow in order to be invariant to both changes in size and to permutations. For the latter, no parameter in the network can be optimized in relation to any specific index in the input vector. In order to be able to take in inputs with varying lengths after optimization, the input must somehow be aggregated to a representation, the size of which does not increase with the number of inputs. An example of such an aggregation is simply to take the average of a range of numbers; regardless of how many numbers there are in the range, their average will always be represented by a single number.

With these requirements in mind, we can choose the simplest solutions to each of them, in order to keep our model as minimal as possible. Thus, the model and its variations presented below do not include a Transformer layer like in the model of Tang and Ha \cite{tang2021sensory}. Indeed, in one ablation study, we show that it is possible to evolve a network that is invariant to permutations of its input vector and to changes in the input size, even when all elements of the network are simple feedforward networks. Kirsch et al. \cite{kirsch2021introducing} aims at meta-learning a black-box reinforcement learning algorithm with a shallow network structure that has invariance properties for both inputs and outputs. In terms of the model structure, our model is similar to that of Kirsch et al., but adding an integrator unit (explained further in Section \ref{section:approach}) gives us the flexibility to choose to focus only on invariance properties for the input. This makes our model comparably easy to optimize. However, we also show how our model can easily be extended to also have invariance properties for the output.



\section{Related Work}
\label{section:related}
Transformer-based models have properties that allow them to take sequences of different lengths as input \cite{vaswani2017attention}. Further, in language tasks where Transformers in recent years have been used to great effect \cite{devlin2018bert, brown2020language}, explicit steps must be taken in order to avoid invariance to permutation of the inputs. This is because it is most often useful to be able to interpret words differently depending on where in a sentence they occur. The use of Transformers in RL-tasks is less frequent (but see \cite{chen2021decision,gupta2022metamorph}). Recently, however, Tang and Ha \cite{tang2021sensory} presented a model to control artificial agents that utilized the properties of Transformers to gain invariance properties for the input. The fact that their Transformer-based model comply with both conditions for permutation and size invariance, can be seen by considering that the key, query, and value transformations use shared parameters for all instances in the input. Then, by fixing the size of one of the transformation matrices ($Q$ in their model), as opposed to letting it depend on the input, the attention matrix is always reduced to a representation of the same size, regardless of the number of elements in the input. Together, this means that parameters in the rest of the network can be optimized in relation to indices in the aggregated representation without being related to any specific indices in the input vector. This highlights relatedness of the problems of size and permutation invariance. By meeting the condition for size invariance by aggregation of the inputs, the problem of input permutation invariance is contained in the part of the network prior to the aggregation. The rest of the network can thus be structured as any normal network. However, attention-based aggregation is not the only type that meets the conditions; as we show below, a simpler aggregation by averaging can also be used.

Plastic neural networks \cite{soltoggio2018born, coleman2012evolving} is an other class of models that under the right circumstances can be permutation invariant as well as size invariant. Although this field can be related to Transformers and Fast Weight Programmers \cite{schlag2021linear}, plastic networks are usually framed quite differently, with a stronger emphasis on biological inspiration. Plastic networks are also more often used for RL-tasks \cite{soltoggio2018born}. A neural network can be considered to be plastic if some function updates the connection strengths of the network during the network's lifetime \cite{mouret2014artificial}.  Not all plastic networks have the properties that we are interested in here. Interestingly, plastic neural networks, where a single plasticity mechanism governs all connections in randomly initialized networks, automatically meet conditions to be invariant to permutations in the input, as well as to changes in size. This observation might give a clue as to how biological neural networks achieve their high level of architectural flexibility. Biological brains learn complicated tasks as a whole \cite{caligiore2019super}, but vary in the number of neurons over a lifetime \cite{breedlove2013biological}. Of course, the brain is governed by a plethora of different plasticity mechanisms \cite{abbott2000synaptic, dan2004spike, dayan2012twenty}, not just a single one, but not all parts of a brain might necessarily need to be invariant in relation to all other parts. Further, the conditions could also be met if there instead of a single plasticity mechanism is a meta-function that organizes local plasticity mechanisms, just as the plasticity mechanisms in turn organize the individual connection strengths.

An early example of a plastic neural network governed by a single learning rule is that of Chalmers \cite{chalmers1991evolution} that evolves a learning rule to update randomly initialized weights of a shallow neural network.
A more recent approach that also falls into this category is presented by Yaman et al. \cite{yaman2021evolving}. In this work, the authors evolve a single discrete Hebbian rule to change the synapses of a randomly initialized network to solve a simple foraging task. They also test their rule's ability to control networks with more hidden neurons. More closely related to our method, is the work of Bertens and Lee \cite{bertens2020network}. They evolve a set of recurrent neural network cells and use them as basic units to form a network between them. In this approach, the synapses and neurons of the overall network are thus recurrent units, and the plastic parameters are the hidden states of the recurrent units. Since all synaptic recurrent units share parameters, and all neural recurrent units share parameters, the network is effectively governed by one overall, homogeneous non-linear platicity mechanism. The network is shown to be permutation invariant, and 
this type of network was shown to be able to solve a simple t-maze task with non-stationary rewards.

Closely related to the approach of Bertens and Lee \cite{bertens2020network} is the work of Kirsch et al.  \cite{kirsch2021introducing}. They evolve a shallow network structure where all connections consist of recurrent neural networks with shared parameters. Like Bertens and Lee, the goal of the approach of Kirsch et al. is to evolve a network with learning capabilities, and they showcase the abilities of their network to exhibit some level of learning in tasks unseen during training. As we extend our model to have invariance properties in the output vector as well as the input in Section ~\ref{subsubsection:output}, we show that we can have an intermediate integrator unit, between input and output units, and is therefore not bound to a shallow network structure like Kirsch et al. \cite{kirsch2021introducing}.  Common to all methods mentioned in this section is that no parameters are optimized in relation to any specific index in the network. Further, new elements can be added to the network architecture after optimization. This is possible as such added elements will be adapted by the same plasticity mechanism as all other elements in the network.

\begin{figure}
\includegraphics[scale=0.105]{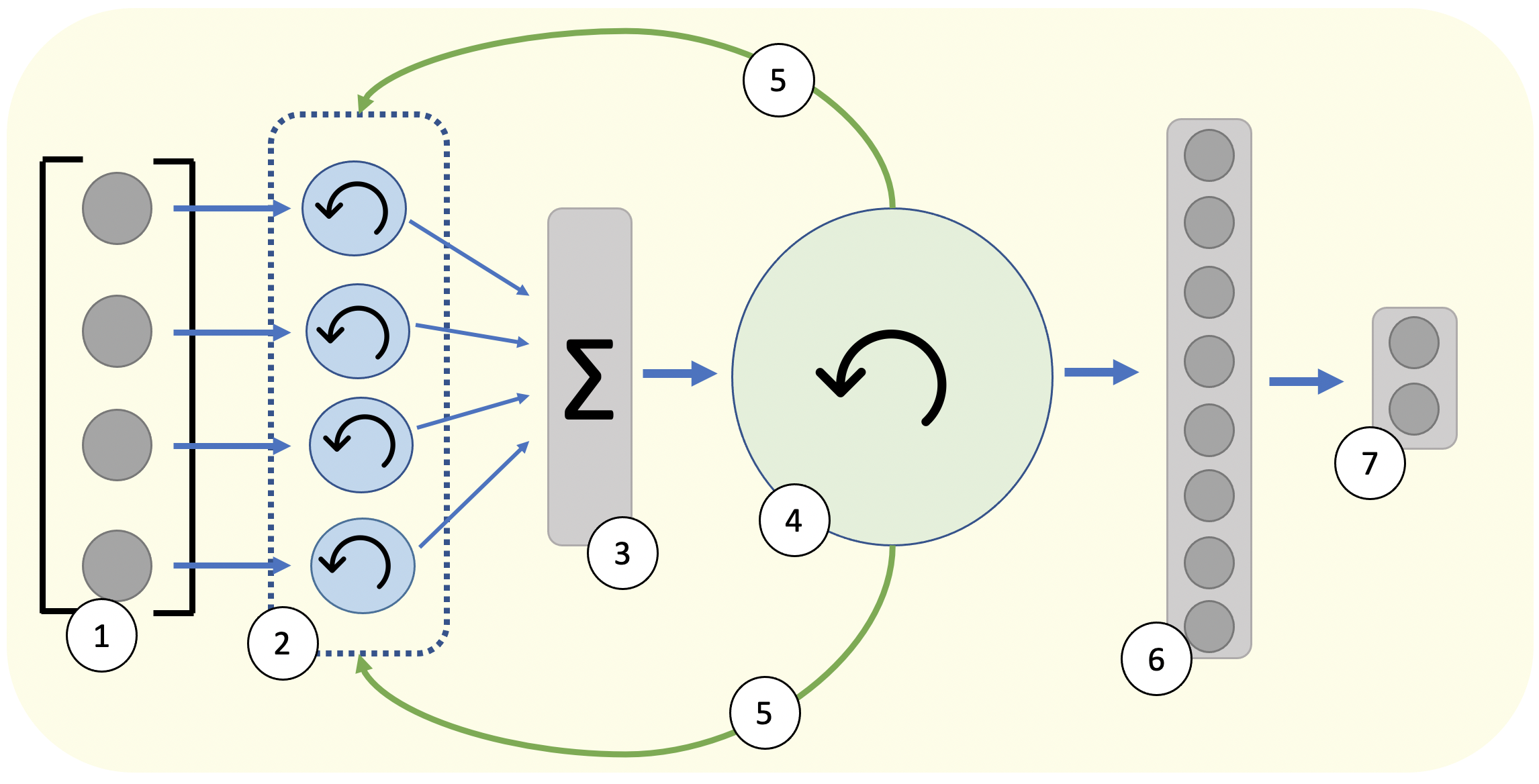}
\caption{Model Overview: \normalfont  At each time step,the external input is presented to the network (1). Each element of the input vector is send into a separate recurrent neural network (RNN) cell that we call \emph{input units} (2). The RNNs in (2) all share the same parameters in their gates. This means that none of the parameters are evolved to be specific to a single element in the external input. The output vectors of the RNNs in (2) are then summed onto a single vector (3). This vector is then passed on to a single RNN (4). The output of this RNN is used to update the hidden states of the RNNs in (2). The same output is also propagated through a dense layer (6) that is connected to the final output of the model (7). }
\centering
\label{fig:overview}
\end{figure}

\section{Approach: A Minimal Neural Model for Permutation Invariance}
\label{section:approach}
The requirement a network needs to meet in order to be invariant to permutations in the input is that no parameter can be optimized in relation to any specific element in the input. This has to hold throughout the entire network. As noted by Tang and Ha \cite{tang2021sensory}, a trivial strategy to achieve this is to constantly permute the input vector throughout the optimization phase. The logic behind this strategy is similar to that of data augmentation strategies used in some image classification studies \cite{taylor2018improving}. If the original dataset does not include examples of rotated objects, a way to make the classifier more general is to augment the dataset with rotated copies of original images. Rotation is a continuous operation and meaningful interpolations can be made between rotations with different angles. However, permutations of indices are discrete and have no meaningful interpolations between them. This means that we would need to augment the dataset with every single possible permutation and thus potentially increase optimization time exponentially.

A visual presentation of the model introduced here is shown in Figure~\ref{fig:overview}. When the input vector from the environment is presented to the model at each time step, each element of the input vector is passed into a separate \emph{input unit}. These \emph{input units} are a type of recurrent neural network called Gated Recurrent Unit (GRU) \cite{cho2014properties} with an added output gate. Other types of RNNs, such as the Long Short-Term Memory unit \cite{hochreiter1997long}, or any of the many other RNN variations \cite{yu2019review} could also have been used.  Importantly, the \emph{input units} all share the same weight matrices. This means that regardless of how many elements there are in the input vector, we only optimize parameters for a single \emph{input unit} and copy these to all the \emph{input units}. 

This separation of the input elements to \emph{input units} with shared parameters is crucial for achieving input permutation invariance, and is a major difference from how inputs are processed by traditional deep RNNs. Note, that even though the \emph{input units} share their evolved weight matrices, their hidden states and their outputs are not necessarily the same at any given time, since these are influenced by the different inputs presented to each input unit. With this approach of routing the input elements into separate units, our method can be described as falling under the category of \emph{instance slot} models (reviewed by Greff, van Steenkiste \& Schmidhuber \cite{greff2020binding}).

Each \emph{input unit} outputs a vector of dimension $(m,1)$. All these vectors are summed into a single vector of dimension $(m,1)$ each element of this vector is divided by the number of \emph{input units}. This averaging of the \emph{input units}' outputs will have the same dimensionality of $(m,1)$ regardless of the number of \emph{input units}. This vector is analogous to the \emph{global latent code} in the approach by Tang and Ha \cite{tang2021sensory}. It is then passed to another RNN (also a GRU with an output gate). We call this the \emph{integrator unit}. It processes the summed outputs of the \emph{input units} just as any normal GRU cell with an output gate would do. The output vector of the \emph{integrator} is passed through two dense layers, the last of which projects to a vector with the number of elements that is needed to make an action in the given environment. The output vector of the \emph{integrator} is also fed back to the \emph{input units}. The \emph{input units} get their hidden states updated through a separate set of GRU gates, the parameters of which are also shared between all the \emph{input units}.

This model complies with both requirements for permutation and size invariance. First, no parameters are specifically optimized in relation to any specific input index. This is ensured by making the \emph{input units} share their optimized parameters, and averaging all their outputs to a single vector. The optimized parameters of the rest of the network are optimized in relation to indices of this aggregate of input units' outputs, but these cannot be traced back to any specific indices of the input vector. Second, the averaging also means that we can add any number of the input units without disrupting the structure of the rest of the network, and without the need for additional optimized parameters.
All RNN cells in all experiments below are Gated Recurrent Units with and additional output gate. The hidden state and the outputs are determined as follows \cite{cho2014properties}:
\begin{equation}
 \mathbf{z}_t = \sigma( W_z[\mathbf{h}_{t-1}, \mathbf{x}_t]+\mathbf{b}_z  ) , 
\end{equation}
\begin{equation}
 \mathbf{r}_t = \sigma( W_r[\mathbf{h}_{t-1}, \mathbf{x}_t]+\mathbf{b}_r  ) , 
\end{equation}
\begin{equation}
 \mathbf{g}_t = tanh( W_g[ \mathbf{r}_t \odot \mathbf{h}_{t-1}, \mathbf{x}_t]+\mathbf{b}_g  ) , 
\end{equation}
\begin{equation}
 \mathbf{h}_t = (1 -\mathbf{z}_t ) \odot \mathbf{h}_{t-1} + \mathbf{z}_t \odot \mathbf{g}_t, 
\end{equation}
\begin{equation}
 \mathbf{o}_t = W_o[\mathbf{h}_{t}, \mathbf{x}_t]+\mathbf{b}_o , 
\end{equation}

Here,$\sigma$ is the sigmoid function, $\odot$ is the Hadamard product, $\mathbf{h}_t$ is the hidden state of the recurrent unit and $\mathbf{o}_t$ is the output of the recurrent unit. $W_i$ and $\mathbf{b}_i$ are weight matrices and bias vectors, respectively.

\todo{describe the symbols in this formula}

\section{Experiments}

\label{subsection:experiments}
We test the model described in Section~\ref{section:approach}, as well as several variations of it with different ablations and three different environments. The particulars of each of these are specified in the sections below. 

\subsection{Environments}
\label{subsection:environments}
\begin{figure}
\includegraphics[scale=0.105]{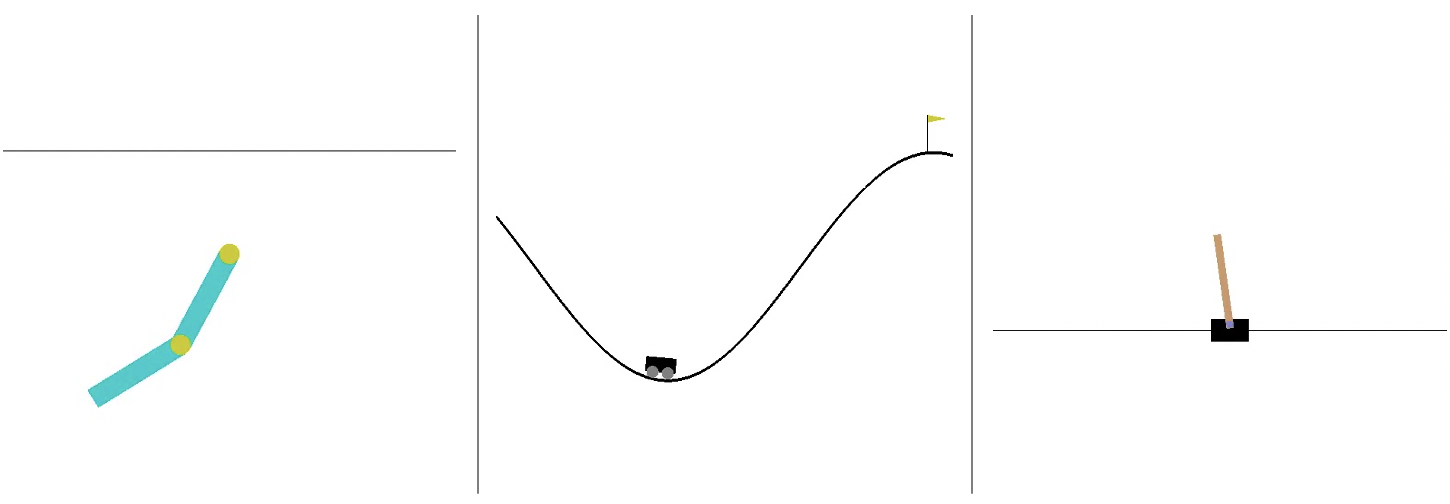}
\caption{Environments Used in Experiments \normalfont From left to right: Acrobot-v1, MountainCar-v0, CartPole-v1}
\centering
\label{fig:environments}
\end{figure}

We test our model in multiple simple control tasks (Fig.~\ref{fig:environments}) from the OpenAI gym suite \cite{brockman2016openai}. In all experiments, the ordering of the inputs to the models stay fixed throughout the entire optimization time.

\subsubsection{CartPole-v1}
\label{subsubsection:cart}
In this classic control task, a pole is balancing on a cart and the agent needs to control the cart such that the pole does not fall for as long as possible \cite{barto1983neuronlike}. The environment has four inputs and two discrete actions. A stopping condition score for the evolution strategy (see below) was set to 495 for this environment.

\subsubsection{Acrobot-v1}
\label{subsubsection:acro}
The task in \emph{Acrobot-v1} is to move a fixed 2-dimensional robotic arm with two joints such that non-fixed end of the arm reaches a certain height \cite{sutton1995generalization}. The faster this is achieved the better the score, and a fitness point of -1 is given for each time step spend. The environment has six inputs and three discrete actions. We set the stopping condition to be an average score of -96.

\subsubsection{MonutainCar-v0}
\label{subsubsection:mount}
The goal is to move a car from a valley on a one-dimensional track up a large hill \cite{moore1990efficient}. The car needs to build up momentum by first moving up a smaller hill on the opposite side. The environment has two inputs and three discrete actions. We set the stopping condition to be an average score of -105.

\subsection{Optimization Details} 
\label{subsection:ES}

We use an evolutionary strategy \cite{salimans2017evolution} (ES) to optimize the parameters of the system. We use an off-the-shelf implementation of ES \cite{ha2017evolving} with its default hyperparameters, except that we set weight decay to zero (unless stated otherwise, hyperparameter configurations of all experiments matches those in Table~\ref{tab:hyper}).  However, for experiments in the Mountain Car experiments, weight decay is set to $0.01$. This is because successes in this environment are initially very sparse, and if all individuals score the same, there is an increased risk of the evolution getting stuck. Weight decay helps evolution differentiating between individuals. The implementation uses mirrored sampling, fitness ranking, and uses the Adam optimizer for optimization. This optimization implementation is similar to that used by Palm et al. \cite{palm2021testing, palm2020} and Pedersen and Risi \cite{pedersen2021evolving}.  Every 20th generation, the mean solution of the population is evaluated over 128 episodes, and if it achieves an acceptable average score, the solution is saved and the evolution run is ended. For all experiments, if a solution was not found within $5,000$ generations, the run was terminated.

\begin{table}
  \caption{Hyperparameters for ES}
  \label{tab:hyper}
  \begin{tabular}{cc}
    \toprule
    Parameter& Value\\
    \midrule
    Population Size & 128\\
    Learning Rate & 0.1\\
    Learning Rate Decay & 0.9999\\
    Learning Rate Limit& 0.001\\
    Sigma & 0.1\\
    Sigma Decay& 0.999\\
    Sigma Limit & 0.01\\
    Weight Decay & 0\\

  \bottomrule
\end{tabular}
\end{table}

\subsection{Model Ablations}

\subsubsection{Full Model}
\label{subsubsection:full}
We refer to the model described in Section~\ref{section:approach} as the full model. It is fully specified by the following weights matrices: Four weight matrices and bias vectors in the \emph{input units} that adjusts the units' hidden states according to the external input following equations (1) through (5) in Section~\ref{section:approach}. Three more weight matrices and bias vectors in the \emph{input units} that adjusts the units' hidden states according to the feedback from the \emph{integrator} following equations (1) through (4). The feedback only serves to adjust the hidden states of the units, and there are thus no output gates for the feedback. Then, there are the four weight matrices and bias vectors in the \emph{integrator}, and two weight matrices and bias vectors for the dense layers that determine the final output of the model. 
The sizes of the matrices of the GRUs are fully determined by the their input sizes, the size of their hidden states and the output sizes. These hyperparameters are summarized in \ref{tab:sizes}. Each \emph{input unit} receives a value from a particular environmental input element, copied eight times to produce a vector. The input size of eight for in the \emph{input units} was chosen partly to dictate the sizes of the following matrices, and to ensure that the input would be able to impact the hidden state better than if it had just been represented by a single value. The sizes of the networks for different tasks only differ in the output vector. CartPole-v1 has two discrete actions, whereas MountainCar-v0 and Acrobot-v1 both have three. For each time step in any environment the index of the largest element of the output vector becomes the action taken by the agent at that time step.
In the beginning of each episode, all hidden states are initialized with noise from a Normal Distribution, $\mathcal{N}(0, 0.05)$. The total number of optimized parameters in the network is $24,064$ for the Cart-Pole environment and $24,096$ for the two other environments.

\begin{table}
  \caption{Network Size Specifications}
  \label{tab:sizes}
  \begin{tabular}{cc}
    \toprule
    Name & Size\\
    \midrule
    Input Unit In & 8\\
    Input Unit Hidden & 16 \\
    Input Unit Out & 24\\
    Inp. Un. Feedback In. & 24 \\
    Integrator In & 24\\
    Integrator Hidden & 16\\
    Integrator Out& 24\\
    Dense 1 & 32\\
    Dense 2 & environment dependent\\
  \bottomrule
\end{tabular}
\end{table}

\subsubsection{No Feedback Model}
\label{subsubsection:nofeed}
We run the same experiments with a variation of the full model with the only difference being that the \emph{input units} do not receive a feedback signal from the \emph{integrator}. The total number of optimized parameters in the network is $5,584$ for the Cart-Pole environment and $5,616$ for the two other environments.

\subsubsection{Integrator as Feedforward Network}
\label{subsubsection:neuffn}
In another variation of the full model, we skip the \emph{integrator} and send the averaged outputs of the \emph{input units} directly to the first dense layer. The \emph{input units} receive the output of the first dense layer as feedback to adjust their hidden states. Total number of optimized parameters for Cart-Pole: $20,904$, others: $20,928$.

\subsubsection{Input Units as Feedforward Networks}
\label{subsubsection:synffn}
In this variation, the \emph{input units} are not GRUs but simple feedforward networks with multiple hidden layers. As in the other variations, the optimized parameters in the weight matrices are shared between the input units. The number of hidden units in the layers of the input units are from beginning to end: 8, 32, 24, 24, 24. The activation function for all the layers is \emph{tanh}. The rest of the network is identical to the full model with no feedback. Total number of optimized parameters for Cart-Pole: $6,064$, others: $6,096$.

\subsubsection{No RNNs}
\label{subsubsection:nornn}
Finally, we run experiments with a model with no recurrence in the network at all; \emph{input units} are feedforward networks with multiple hidden layers and their averaged outputs are send through multiple dense layers. The size of the \emph{input units} are the same as in Section~\ref{subsubsection:synffn}, and the average output is then send through layers of sizes: 24, 32, 24, 24, 16, 32, before being projected to the number of actions in the given environment. Total number of optimized parameters for Cart-Pole: $5,760$, others: $5,792$.


\subsubsection{Standard RNN}
\label{subsubsection:rnn}
In addition to variations of our proposed model, we run experiments with a traditional RNN structure. This consists a GRU unit with additional output gates following equations (1) through (5) in Section ~\ref{section:approach}. The input size of the this RNN is equal to the input vector given by the environment that it is optimized in. Its hidden state has 16 elements, and so does its output. It is connected to two dense layers, one with 32 hidden notes, and one that has a number of nodes equal to the number of possible actions in the environment. Total number of optimized parameters for Cart-Pole: $1,954$, Mountain Car: $1,859$, Acrobot:$2,115$.

\subsubsection{Output Permutations}
\label{subsubsection:output}
In our last experiment, we adapt the full model to also be invariant to permutations and changes in size to the output of the network.
The first half of this model is identical the that of the full model. However, instead of projecting to a dense layer, the \emph{integrator} projects to two units of the same type as the \emph{input units,} but with their own set of shared optimized parameters. In this experiment, these \emph{output units} have weight matrices of sizes equal to those of the \emph{input units}, but this is not required. The \emph{output units} have no weight matrices for feedback. Only experiments on the Cart-Pole environment is done with this model and the total number of optimized parameters is $24,176$. For these evolution runs, weight decay was set to $0.01$. Further, the fitness of each individual of a generation was here the average performance over four episodes, instead of just a single episode. We extend the optimization is this way, as we are now attempting to solve another problem on top of what was solved by the previous models.

\begin{figure}
\includegraphics[scale=0.105]{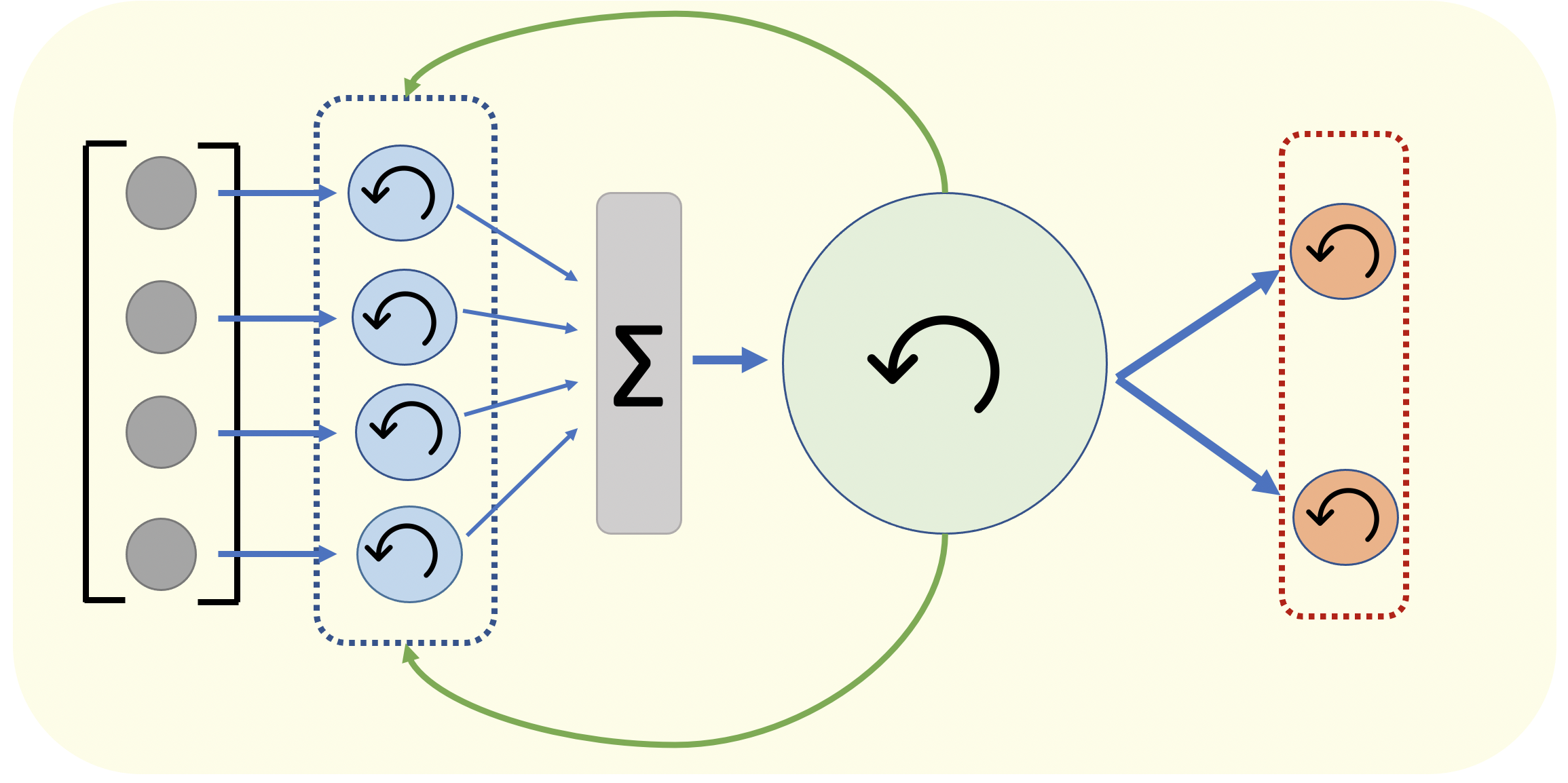}
\caption{Model Extended to Output Permutation Invariance: \normalfont  The first part of this model is identical to that in Fig.~\ref{fig:overview}. However, instead of dense layers projecting to the output, this model has two recurrent units with shared parameters as the final output nodes. }
\centering
\label{fig:out_overview}
\end{figure}

\section{Results}
\label{section:results}

\begin{figure*} 
\includegraphics[scale=0.38]{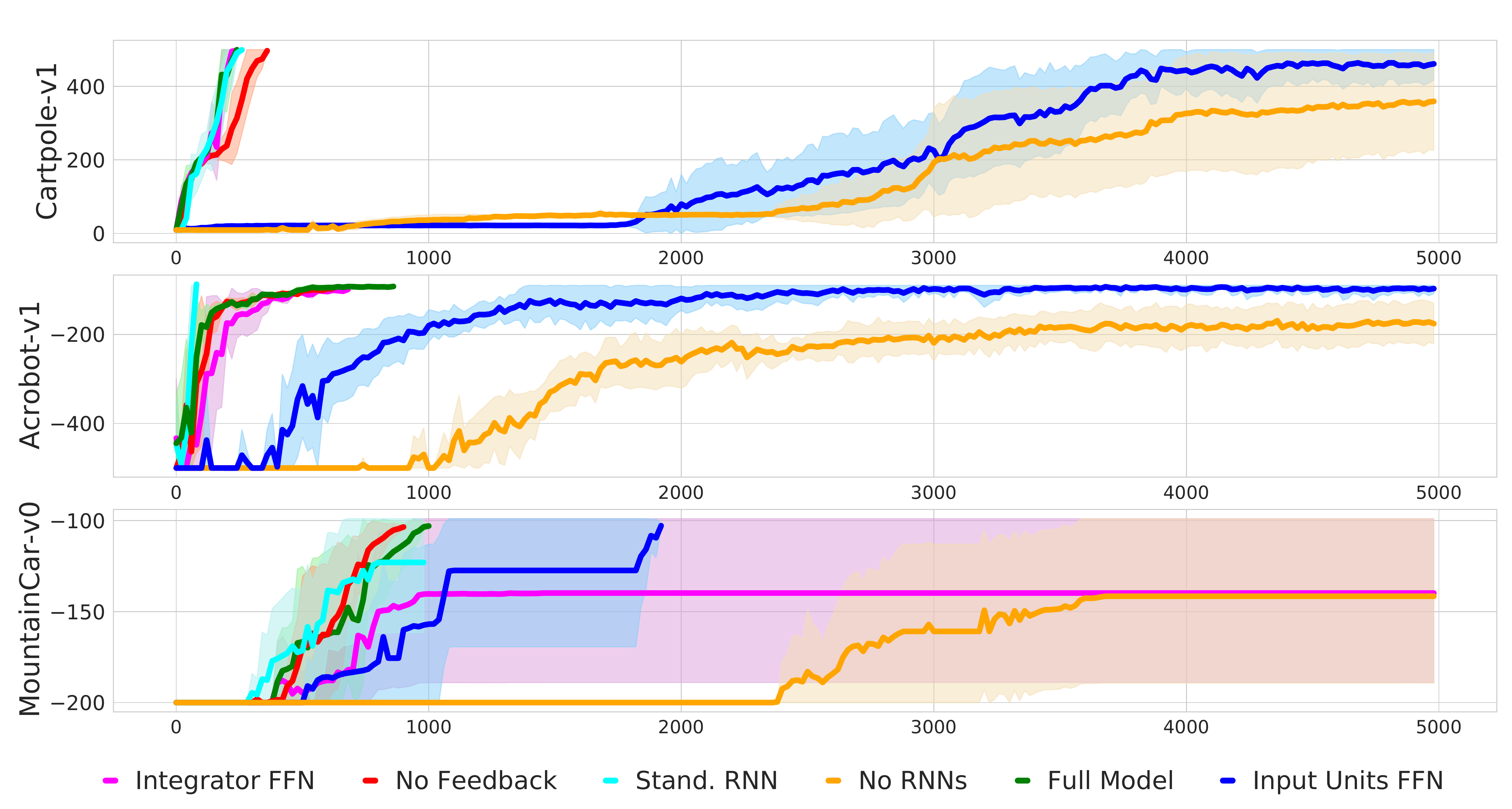}
\caption{Training Curves: \normalfont Means and standard deviations. Each curve represents the mean of five independent evolution runs with a specific method. The full model as described in Figure \ref{fig:overview} and its variation without feedback from the integrator to the \emph{input units} tend to find solutions to the tasks quickly. The same is true for the standard RNN. When the \emph{input units} are feedforward networks rather than RNNs, evolution in most cases did not end up finding a solution within the set time limit of 5000 generations. }
\label{fig:aggr}
\end{figure*}

\begin{figure}
\includegraphics[scale=0.18]{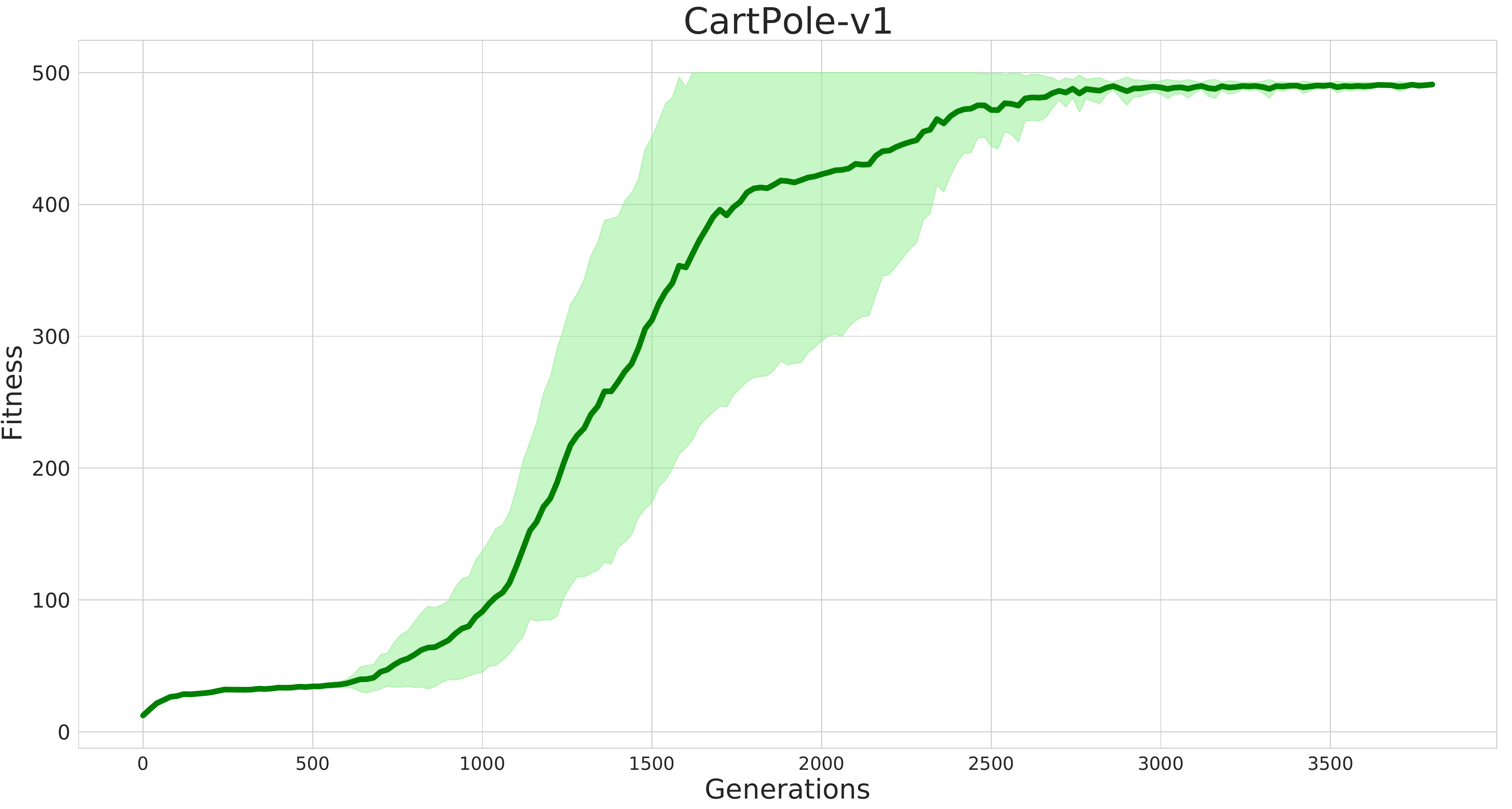}
\captionsetup{justification=centering}
\caption{Training Curve for Model with Output Units.  \normalfont Means and standard variations of five independent evolution runs.  }
\centering
\label{fig:cart_outperm}
\end{figure}

Training curves of each experiment are shown in Figure \ref{fig:aggr}, except for the experiment described in Section ~\ref{subsubsection:output} that is presented in Figure \ref{fig:cart_outperm}. Each curve represents the mean of five independent evolutionary runs. From Figure~\ref{fig:aggr}, it is clear that evolution tends to find solutions much faster with some models than for others. Specifically, the full model, the full model without feedback and the standard RNN finds solutions in the matter of hundreds of generations for all problems. On the contrary, for the models where the \emph{input units} are not RNNs a solution was often not found within the time limit of 5,000 generations.

After optimization, we are interested in how well the models do with online permutations of the input vectors. In Table ~\ref{tab:eval}, such evaluations are shown for an optimized full model, the model consisting of feedforward units only, and the standard RNN model. Across the board, the models designed for invariance tend to achieve similar scores under all conditions, even when the inputs are shuffled at intervals as frequent as every 5th time step. The standard RNN fails under all permutation conditions. Following \cite{tang2021sensory}, we also evaluate the models when given a larger input vector than seen during optimization with redundant values. Here too, the models designed for invariance show no signs of deterioration in performance. It was not possible to evaluate the standard RNN using a doubled input size, due to its rigid structure.

\begin{figure}
\includegraphics[scale=0.18]{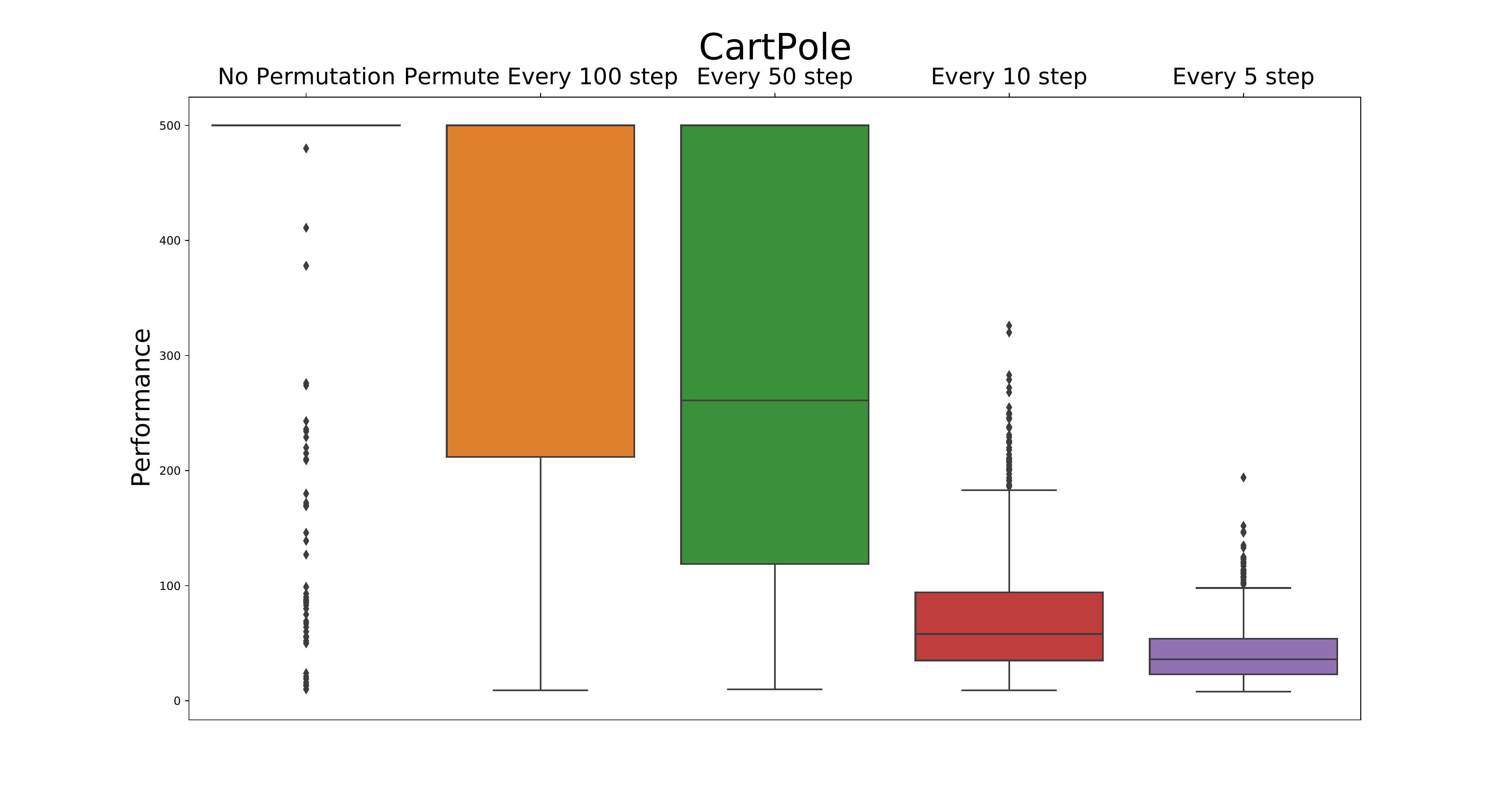}
\captionsetup{justification=centering}
\caption{Performance Under Online Permutation of Input and Output. \normalfont The model performs well under random permutations of both the input and output when the random ordering is fixed during the episode. However, online permutations makes the model fail at increasing levels. }
\label{fig:out_perm_results}
\end{figure}

Figure ~\ref{fig:cart_outperm} shows that it takes longer for evolution to find solutions to the Cart-Pole environment when the full model is extended to also have \emph{output units} with shared parameters, but that solutions are consistently found. Further, in Figure ~\ref{fig:out_perm_results}, we see that the optimized model does not tend to perform well with frequent permutations of both the input and output vectors. However, as shown by the left-most box plot in the figure, the model performs well under random orderings, as long as they stay fixed during the episode.

\begin{figure*} 
\includegraphics[scale=0.31]{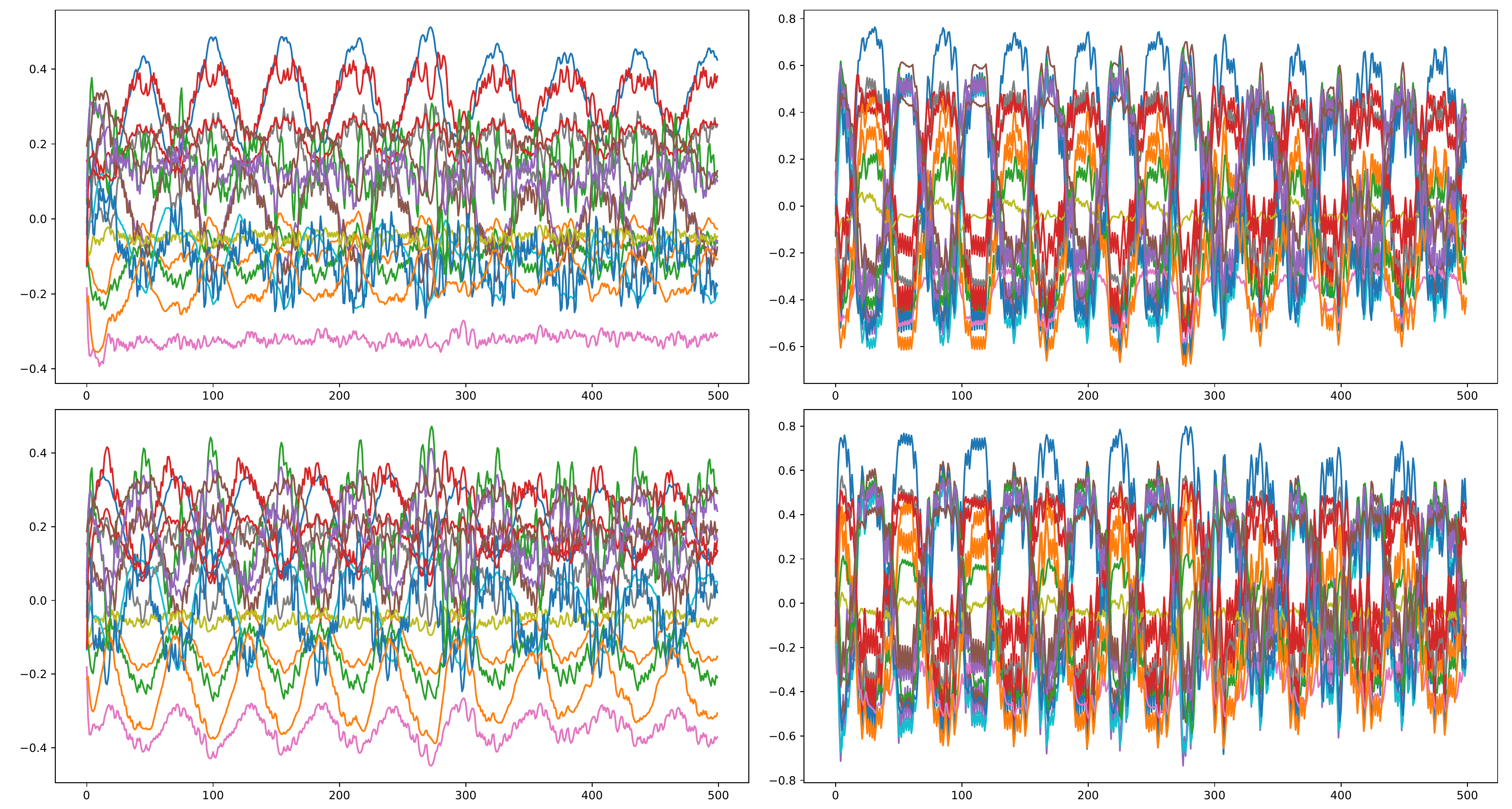}
\caption{CartPole Hidden States: No Shuffling \normalfont The 16 hidden state elements of each of the four input unit over a full episode in the Cart-Pole environment. The units seem to have seperate, fixed roles.}
\centering
\label{fig:noshuffle}
\end{figure*}

\begin{figure*} 
\includegraphics[scale=0.31]{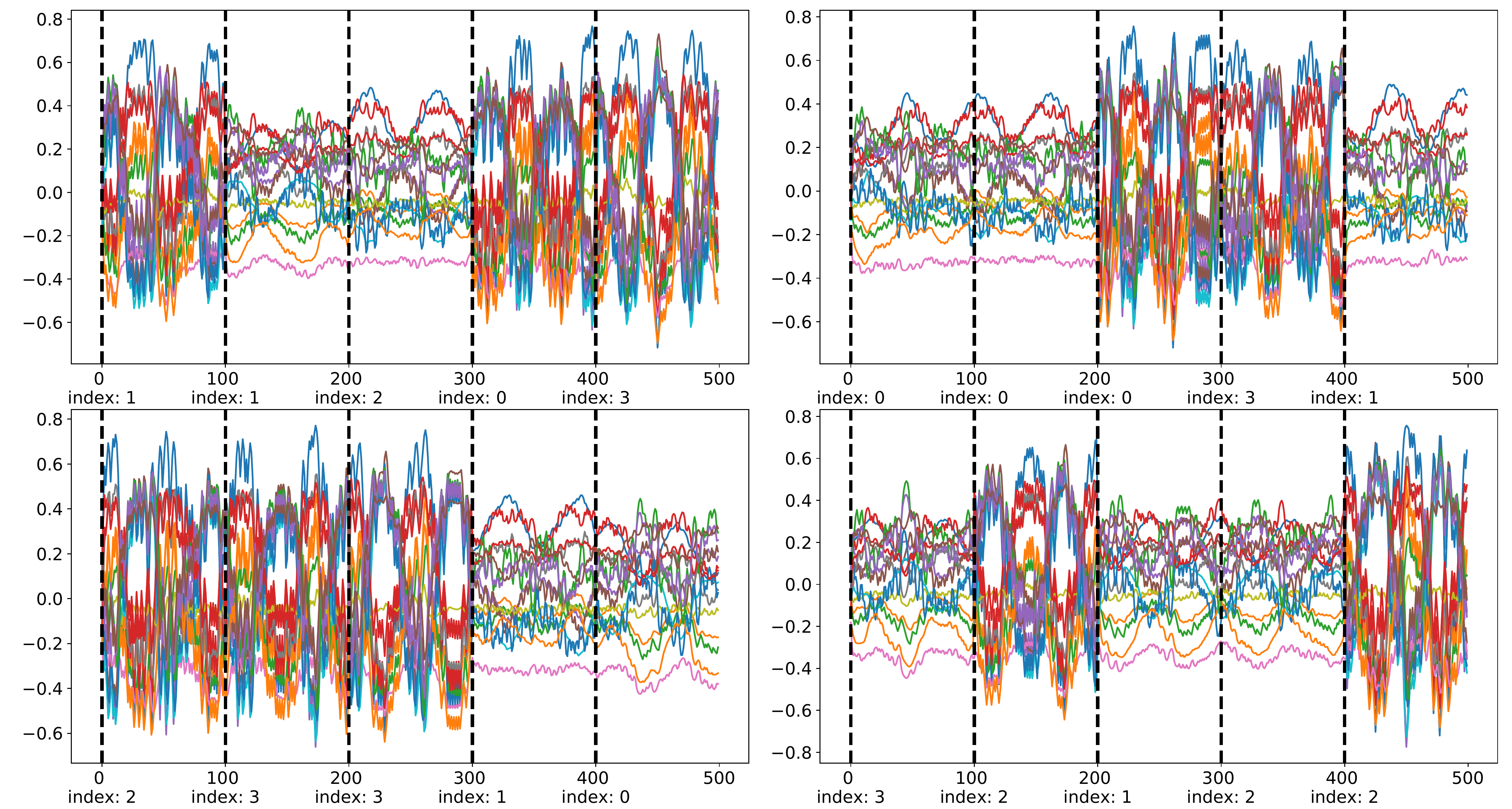}
\caption{CartPole Hidden States: Shuffle Every 100th Step \normalfont Black dotted lines indicate the times at which the input vector was randomly permuted. The hidden states rapidly adapt their activity levels in response to permutations.}
\centering
\label{fig:shuffle}
\end{figure*}

We can look closer at how the \emph{input units} behave under conditions without and with online input permutations. Such cases are presented for a full model performing in the Cart-Pole environment in Figure ~\ref{fig:noshuffle} and Figure ~\ref{fig:shuffle} respectively, where we see the 16 hidden state elements of each of the four \emph{input units} over a full episode. Figure ~\ref{fig:noshuffle} shows that when no permutations occur, the mode of each \emph{input unit} tends to look similar throughout the episode. However, as can be seen in Figure ~\ref{fig:shuffle}, the \emph{input units} are able to quickly switch roles in response to a permutation.

\begin{table*}
  \caption{Table of Results. \normalfont Means and standard deviations over $1000$ episodes. For each method, we choose the run with the highest population mean score at the end of evolution. Input Doubling means that each element of the input vector is copied. E.g., in the Cart-Pole environment, this means that there are eight input elements and therefore also eight \emph{input units} instead of four. No. Perm. means that the input vector is not permuted online. However, in the beginning of each new episode, the ordering is randomized. Results show that variations of our model both with and without recurrent dynamics are able to do well in the tasks, even when the input is permuted online several times. Note, however, as seen in Figure \ref{fig:aggr}, evolution runs of the model without recurrent dynamics, did not reliably result in a solution within the set time limit. The standard RNN model does not do well in any of these scenarios.}
  \label{tab:eval}
  \begin{tabular}{ccccccc}
    \toprule
    Full Model \\ \hline
    Env. & Input Doubling & No Perm. & Every 100 & Every 50 & Every 10 & Every 5\\
    \midrule
    Cart-Pole & 488.3 $\pm$ 67.9 & 485.2 $\pm$ 76.7 & 488.5 $\pm$ 68.2 & 488.9 $\pm$ 68.4 & 489.6 $\pm$ 66.0 & 481.8 $\pm$ 86.7\\
    Acrobot & -106.3 $\pm$ 48.2 & -108.9 $\pm$ 57.9 & -106.8 $\pm$54.9 & -107.1 $\pm$ 57.1 & -108.2 $\pm$ 60.7 & -104.5 $\pm$ 44.7 \\
    Mountain Car & -99.7 $\pm$ 5.6 & -99.5 $\pm$ 5.7 & -99.7 $\pm$ 5.5 & -99.9 $\pm$ 5.8 & -100.0 $\pm$ 5.6 & -107.5 $\pm$ 21.6\\  \hline
    Standard RNN \\ \hline
    Cart-Pole & N/A & 172.8 $\pm$ 210.0 & 73.4 $\pm$ 87.3 & 47.8  $\pm$ 50.7 & 20.1  $\pm$ 14.5 & 23.1  $\pm$ 13.4 \\
    Acrobot & N/A & -396.3 $\pm$ 168.4 & -293.4 $\pm$ 152.1 & -295.9 $\pm$ 137.6 & -279.3 $\pm$ 108.0 & -280.1 $\pm$ 106.1 \\
    Mountain Car & N/A & -149.4 $\pm$ 48.3 & -146.3 $\pm$ 45.8 & -171.4 $\pm$ 40.8 & -180.1 $\pm$  34.7 & -193.3$\pm$ 20.9 \\ \hline
    No RNNs \\ \hline
    Cart-Pole & 496.4 $\pm$ 35.4  & 496.4 $\pm$ 35.7 & 494.7 $\pm$ 43.4 & 496.3 $\pm$ 36.5 & 497.6 $\pm$ 29.1 & 496.1$\pm$37.2 \\
    Acrobot & -118.7$\pm$ 63.0  & -118.5$\pm$ 63.3 & -121.8$\pm$68.5 & -120.1$\pm$70.1 & -123.9$\pm$76.2 & -116.5$\pm$ 61.6 \\
    Mountain Car & -105.5 $\pm$ 7.0 & -104.8 $\pm$ 7.6 & -105.2 $\pm$ 7.0 & -105.0$\pm$7.0 & -105.2$\pm$7.7 & -105.0$\pm$7.4 \\
  \bottomrule

\end{tabular}
\end{table*}

\section{Discussion}
\label{section:discussion}
In this paper we demonstrate that the requirements needed for making a network invariant to permutation and size changes of the external inputs can be met by relatively simple models. Importantly, no parameters can be optimized in relation to any specific index of the input vector, and projections of the input must at some point in the network be aggregated to a representation that does not grow with the number of inputs. 

The simple solution we use in the model presented here, is to average the outputs of input units with shared parameters. Even when optimized with fixed inputs, the models are remarkably robust to frequent permutations of the input vector. The solution does not specify that the \emph{input units} need to be recurrent neural networks. Indeed, we find that in the environments we use for experimentation here, it is possible to find solutions solely using feedforward networks. However, such solutions tends to be more difficult to find compared to when \emph{input units} are RNNs. This effect might only be exacerbated in more difficult environments. The use of feedback from the \emph{integrator} did not tend to make a difference in our experiments. This could be due to the simplicity of the environments, as Tang and Ha \cite{tang2021sensory} report that their analogous \emph{input units} need to get the model's previous outputs as additional inputs in order to work. One might expect that the more overlapping the values of the input vector can be, the more need there is for some form of global signal as well as a memory of previous inputs.

We further show that it is simple to extend the model to also be able to work with different permutations of the output vector. This is done by following the same principle as for the input vector: no parameters in the network can optimized in relation to a specific index of the output vector. We solve this by having \emph{output units} with shared parameters that receive a common input from the \emph{integrator}. However, the problem of invariance to permutations of the output vector is different in important ways. First of all, the \emph{output units} almost certainly need to be recurrent, as having different hidden states is the only way that the units are be able to give different outputs in response to the identical inputs they are presented with from the \emph{integrator}. Second, dealing with online permutations of the output vector is much harder than of the input vector as indicated by the results in Figure ~\ref{fig:out_perm_results}. This is not surprising considering that every time the output vector is permuted, the next action of the network will be random. For the Cart-Pole environment with only two actions that are oppositely directed, this might be somewhat feasible, as only a single random action is needed in order to have a perfect overview over all actions. For environments with larger numbers of available actions, the agent would have to behave randomly for potentially many time steps before being able to settle into a learned behavior. Still, even though rapid online permutations of the output might insurmountable at large scales, the properties of a model like the one presented in Section ~\ref{subsubsection:output} can still be interesting. In this model, the number of parameters to be optimized is completely decoupled from the size of the external input and the number of actions in the environment. As such, the model can potentially be optimized on multiple different environments that do not need to share the input and output spaces. This idea is not unlike the one presented by Kirsch et al. \cite{kirsch2021introducing}. However, Kirsch et al. are in their work aiming for evolving a black-box reinforcement learning algorithm. While the optimization procedure of the model presented in this paper could be altered to mimick that of Kirsch et al., our model currently does not take any rewards into account, but focuses solely on solving the problems of invariance. With the model presented here, we get these invariance properties with only a small fraction of the optimization time reported by Kirsch et al. on the same problems.

 \subsection{Future Directions} \label{future}
Having shown that our model can be reliably evolved to be invariant to permutations on simple problems, it is worth considering how the model might be scaled up to bigger problems. In future experiments, we aim at also using our model to solve continuous control problems with more inputs and outputs. Trivially, it is always a possibility to make the model more expressive by increasing the sizes of the weight matrices throughout the network. However, there are other ways the model can be expanded, while still conforming to the laid out restriction. It should for example be possible to add a layer of \emph{input units} parallel to the ones in the model in Figure ~\ref{fig:overview}. Each element in the input vector would thus be send through two different \emph{input units}. The added \emph{input units} would still have to share optimized parameters with each other, but, importantly, not with the "original" layer of \emph{input units}. The averaged output vectors of the \emph{input unit} layers can then be concatenated and send through the network as in Figure ~\ref{fig:overview}. An added, separate set of optimized parameters for processing the input could allow for more specialization, without hurting the ability to deal with permutations of the input.

\section{Conclusion}
\label{section:conclusion}
The world is a messy place and it might not always be possible for us to fully anticipate how inputs will be presented to agents meant to perform in it. With this paper, we contribute to the efforts of making artificial agents more adaptable to changes to their inputs. We do so by being explicit about what the models at the very least will need for being able to adapt to permutations and size changes, and use these restrictions to develop simple models that adhere to them. We hope that this will inspire even more research in making more adaptable artificial agents.

\begin{acks}
 This project was funded by a DFF-Research Project1 grant (9131-00042B). 
\end{acks}

\bibliographystyle{ACM-Reference-Format}
\bibliography{refs}


\end{document}